\documentclass[oneside,11pt]{article}
\usepackage{graphicx}
\usepackage{hyperref}

\begin{document}

\title{An Open Source Pattern Recognition Toolbox for MATLAB}

\author{
  Kenneth D. Morton, Jr.\\
  CoVar Applied Technologies\\
  \texttt{kenny@covartech.com}
  \and
  Peter Torrione\\
  CoVar Applied Technologies\\
  \texttt{pete@covartech.com}
  \and
  Leslie Collins \\
  Duke University \\
  \texttt{lcollins@ee.duke.edu}
  \and
  Sam Keene \\
  The Cooper Union \\
  \texttt{keene@cooper.edu}
}

\maketitle

\begin{abstract}%
Pattern recognition and machine learning are becoming integral parts of algorithms in a wide range of applications. Different algorithms and approaches for machine learning include different tradeoffs between performance and computation, so during algorithm development it is often necessary to explore a variety of different approaches to a given task. A toolbox with a unified framework across multiple pattern recognition techniques enables algorithm developers the ability to rapidly evaluate different choices prior to deployment. MATLAB is a widely used environment for algorithm development and prototyping, and although several MATLAB toolboxes for pattern recognition are currently available these are either incomplete, expensive, or restrictively licensed. In this work we describe a MATLAB toolbox for pattern recognition and machine learning known as the PRT (Pattern Recognition Toolbox), licensed under the permissive MIT license. The PRT includes many popular techniques for data preprocessing, supervised learning, clustering, regression and feature selection, as well as a methodology for combining these components using a simple, uniform syntax. The resulting algorithms can be evaluated using cross-validation and a variety of scoring metrics to ensure robust performance when the algorithm is deployed. This paper presents an overview of the PRT as well as an example of usage on Fisher's Iris dataset.
\end{abstract}


\section[Introduction]{Introduction}
\label{intro}

In this work we describe the PRT, an object oriented framework for pattern recognition within MATLAB developed to enable rapid data and algorithm exploration and enable practitioners to build complex machine learning techniques. The PRT is freely available and released under the MIT license.  The PRT defines standard interfaces for machine learning datasets (prtDataSets) and pattern recognition and machine learning tasks (prtActions). Since prtActions always provide a prtDataSet as an output, individual machine learning actions can be easily combined together to form a machine learning algorithm (a prtAlgorithm). Since prtActions and prtAlgorithms provide a unified calling syntax for cross-validation and algorithm evaluation, algorithmic changes can be rapidly evaluated with few code modifications. Furthermore, the PRT provides extensive support for visualization of results, and provides an easy path for implementing new features.

The PRT also provides a number of other benefits which space restrictions prohibit us from discussing in detail, but which are well \href{http://newfolder.github.io/prtdoc/prtDocUsersGuide.html}{documented}.  These include, k-folds and cross-validation techniques, data visualization methods, a wide array of standard classification and regression techniques (e.g., SVM\cite{DBLP:journals/ml/CortesV95}, RVM \cite{Tipping2001}, Random Forest \cite{DBLP:journals/ml/Breiman01}, PLSDA \cite{DeJong1993}) as well as various pre-processing, feature selection, decision rules, and a suite of tools for modeling data distributions.

The remainder of this paper is organized as follows. Section \ref{keyFeatures}, discusses some of the more important and novel features of the PRT. Section \ref{example} shows a simple example illustrating the usage of the PRT. Section \ref{conclusion} offers some final points.

\section[Key Features] {Key Features}
\label{keyFeatures}
In the PRT features $\mathbf{X}$ and (optionally) targets $\mathbf{Y}$ are contained in and managed by the prtDataSet class, with subclasses that specify the nature of the targets or labels. For example, datasets that utilize integer labels typically used for classification are known as prtDataSetClass objects and those that have real valued targets (as in regression) are known as prtDataSetRegress objects.

In contrast algorithms, their parameters, and the techniques by which they are trained and evaluated are defined and managed by the class prtAction. Example prtActions include classifiers, data preprocessors, feature selection techniques and regression techniques. prtActions provide an interface including two methods: \textbf{train} and \textbf{run}. \textbf{train} takes a prtDataSet and outputs a prtAction of the same type with inferred parameters. \textbf{run}, on the other hand, maps a prtDataSet to another prtDataSet. Since all \textbf{run} methods output prtDataSets, the output of any \textbf{run} method can be used as the input to another prtAction's \textbf{train} method. Therefore, prtActions can be combined together in complicated structures to enable rapid development of multi-stage algorithms.

In the parlance of the PRT, a machine learning \emph{algorithm} is comprised of several individual actions that operate sequentially or in parallel. For example, it is common to first perform data preprocessing followed by classification. In the PRT a prtAlgorithm is a subclass of prtAction which stores and manages a collection of prtActions and the connections between them. Because a prtAlgorithm is also a prtAction, methods such as train, run and kfolds operate on prtAlgorithms as they operate on prtActions.

A prtAlgorithm is constructed by using the $+$ and $/$ operators to combine prtAction objects. The $+$ operator is used for sequential algorithm flow while the $/$ operator is used for parallel algorithm flow. Using these operators complex algorithm flows can be constructed to perform tasks such as classifier fusion. The following section illustrates the construction and use of prtAlgorithms to combine pre-processing and classification actions.

\section[Example Use]{Example use of the PRT}
\label{example}

This section illustrates example use of the PRT using Fisher's Iris dataset \cite{Fisher1936}. We will demonstrate how the PRT can be used to build, evaluate, and visualize machine learning algorithms and data.

A function to generate a prtDataSetClass containing Fisher's iris data is included in the PRT, prtDataGenIris. Fisher's Iris dataset contains $4$ features, the length and width of the sepal and petal, from $3$ different species of iris. This example will focus on binary classification (or detection) of one of these species, setosa, from the other two species. To create a binary classification dataset, a new prtDataSet, \emph{ds}, is created that has the same observations but has binary targets that indicate if an observation is of the setosa class.

For visualization, we elect to project the iris data onto its $2$ dimensional principal components space. Prior to applying principal components it is customary to remove the mean and perform standard deviation normalization of each feature dimension. This can be done by using the prtAction, prtPreProcZmuv (zero mean, unit variance). Following this, principal components analysis can be applied to the data using prtPreProcPca.

Given this dataset, any number of classifiers can be constructed and trained to distinguish between the two classes. We will examine two classifiers, probabilistic maximum \textit{a posteriori} classification (prtClassMap), and the relevance vector machine, (prtClassRvm). Because each classifier is a prtAction, syntax is the same for both classifiers. The following code illustrates how the normalization, PCA and classification operations are easily combined.

\begin{verbatim}
algoMap = prtPreProcZmuv + prtPreProcPca('nComponents',2) + prtClassMap;
algoRvm = prtPreProcZmuv + prtPreProcPca('nComponents',2) + prtClassRvm;
\end{verbatim}

Both algorithms can then be evaluated using $5$ fold random cross-validation using the kfolds method.

\begin{verbatim}
crossValidatedOutputMap = kfolds(algoMap,ds,5);
crossValidatedOutputRvm = kfolds(algoRvm,ds,5);
\end{verbatim}

From the outputs of the above commands the receiver operator characteristics can be plotted to compare the expected performance of the two classifiers as a function of potential thresholds.

\begin{verbatim}
[pfMap, pdMap] = prtScoreRoc(crossValidatedOutputMap);
[pfRvm, pdRvm] = prtScoreRoc(crossValidatedOutputRvm);
\end{verbatim}

\begin{figure}[ht]
\centering
\includegraphics[scale=.4]{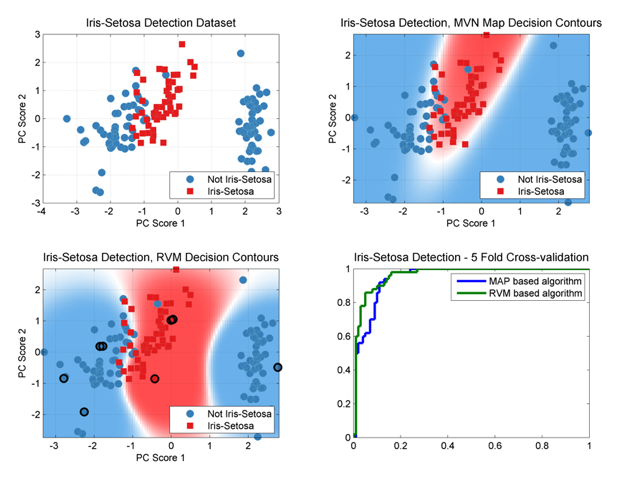}
\caption[Output Image]{A) Visualization of principal components projection. B) Decision contours from maximum \textit{a posteriori} classification. C) Decision contours from using a relevance vector machine.D) Receiver operator characteristics for the detection of the iris setosa.}
\label{fig}
\end{figure}

Figure \ref{fig} shows the results of projecting the data into 2 dimensions, the decision contours that result from both classifiers, and their corresponding receiver operating curves. For brevity, the plotting commands are omitted.

Although this example has focused on a fairly simple and well studied classification problem, several of the most useful aspects of the PRT have been highlighted. The ability to quickly construct prtAlgorithms that share a unified syntax combined with the ability to visualize datasets and classifier decision contours and calculate cross-validated performance metrics enables rapid algorithm design. During the design process a variety of techniques and algorithm flows can be explored and evaluated in the unified framework provided by the PRT.

\section[Conclusion]{Conclusion}
\label{conclusion}

This document has discussed the PRT - an open source and permissively licensed pattern recognition toolbox for MATLAB. The PRT provides a framework for and implementation of many standard machine learning techniques and most importantly it provides methodology for combining individual techniques to form pattern classification algorithms that can be rapidly modified and evaluated. Using the PRT a wide variety of algorithmic possibilities can be explored in a short amount of time and robust operation can be ensured by taking advantage of built-in algorithm cross-validation.

The PRT includes full \href{http://newfolder.github.io/prtdoc/}{documentation} of all functions, a quick start guide, and a unit test suite for all functionality. In addition, the developers maintain an active blog and discussion forum for support of the tool at the following URL: \url{http://newfolder.github.io}. It is compatible with MATLAB version 2008a and later, and therefore runs on Windows, Linux/Unix and Mac platforms. As MATLAB is typically available campus wide at most academic institutions, and is widely used in industry, we feel this tool will be extremely useful to both researchers and practitioners worldwide.

The PRT provides a straightforward \href{http://newfolder.github.io/prtdoc/prtDocAddYourOwn.html}{framework} for algorithm and classifier evaluation, which enables researchers to implement their own algorithms and immediately compare them to standard approaches.  The open-source and permissively licensed nature of the PRT enables other researchers and practitioners to expand the toolbox's capabilities. In addition, template objects are provided that allow users to define their own methods without requiring the use of learning a complicated API. As the product is hosted on the open source hosting site  \href{https://github.com/newfolder/PRT}{GitHub}, the authors would like to encourage any and all contributions.

%
\bibliographystyle{plain}
\bibliography{mortonPrt}

\end{document}